\documentclass{llncs}
\usepackage{makeidx}  

\usepackage{array}  
\usepackage{booktabs}  
\usepackage{ragged2e}  
\usepackage{graphicx}  
\usepackage{color}   
\usepackage{pifont}  
\usepackage{url}     

\newcolumntype{R}[1]{>{\RaggedLeft\arraybackslash}p{#1}}

\begin{document}
\frontmatter          
\pagestyle{headings}  
\addtocmark{Hamiltonian Mechanics} 
\mainmatter              
\title{Conformal Predictors for Compound Activity Prediction}
\titlerunning{Conformal Predictors for Compound Activity Prediction}  
%
\author{Paolo Toccaceli, Ilia Nouretdinov and Alexander Gammerman}
\authorrunning{Paolo Toccaceli et al.} 
%
\tocauthor{Paolo Toccaceli, Ilia Nouretdinov, Alexander Gammerman}
\institute{Royal Holloway, University of London, Egham, United Kingdom,\\
\email{{(paolo, ilia, alex)}@cs.rhul.ac.uk}\\WWW home page: \texttt{http://clrc.rhul.ac.uk/}}

\maketitle              

\begin{abstract}
The paper presents an application of Conformal Predictors to a chemoinformatics problem of identifying activities of chemical compounds. The paper addresses some specific challenges of this domain: a large number of compounds (training examples), high-dimensionality of feature space, sparseness and a strong class imbalance.  A variant of conformal predictors called Inductive Mondrian Conformal Predictor is applied to deal with these challenges. 
Results are presented for several non-conformity measures (NCM) extracted from underlying algorithms and different kernels.   
A number of performance measures are used in order to demonstrate the flexibility of Inductive Mondrian Conformal Predictors in dealing with such a complex set of data.
\end{abstract}
\section{Introduction}
Compound Activity Prediction is one of the key research areas of Chemoinformatics. It is of critical interest for the pharmaceutical industry, as it promises to cut down the costs of the initial screening of compounds by reducing the number of lab tests needed to identify a bioactive compound.
The focus is on providing a set of potentially active compounds that is significantly ``enriched'' in terms of prevalence of bioactive compounds compared to a purely random sample of the compounds under consideration. The paper is an extension of our work presented in~\cite{excape}.

While it is true that this objective in itself could be helped 
with the 
classical machine learning techniques that usually provide a bare 
prediction, the hedged predictions 
made by Conformal Predictors (CP) provide some additional information that can be used advantageously in a number of respects.

First, CPs will supply the valid measures of confidence in the prediction of bioactivities of the compounds.
Second, they can provide prediction and confidence for individual compounds.
Third, they can allow the ranking of compounds to optimize the experimental testing of given samples.  
Finally, the user can control the number of errors and other performance measures like precision and recall by setting up a required level of confidence in the prediction.

\section{Machine learning background}
\subsection{Conformal Predictors}\label{cp01}

Conformal Predictors described in~\cite{algoLearning,HedgingPreds} revolves around the notion of Conformity (or rather of Non-Conformity).

Intuitively, one way of viewing the problem of classification is to assign a label $\hat y$ to a new object $x$ so that the example $(x,\hat y)$ does not look out of place among the training examples $(x_1, y_1),(x_2, y_2),\dots,(x_\ell, y_\ell)$. To find how ``strange'' the new example is in comparison with the training set, we use the Non-Conformity Measure (NCM) to measure  $(x,\hat y)$.

The advantage of approaching classification in this way is that this leads to a novel way to quantify the uncertainty of the prediction, under some rather general hypotheses.

A Non-Conformity Measure can be extracted from any machine learning algorithm, although there is no universal method to choose it. 
Note that we are not necessarily interested in the actual classification resulting from such ``underlying'' machine learning algorithm. What we are really interested in is an indication of how ``unusual'' an example appears, given a training set.

Armed with an NCM, it is possible to compute for any example $(x,y)$ a \textit{$p$-value} that reflects how good the new example from the test set fits  (or conforms) with the training set.
A more accurate and formal statement is:  for a chosen $\epsilon \in [0,1]$ it is possible to compute  $p$-values for test objects so that they are (in the long run) smaller than $\epsilon$ with probability at most $\epsilon$.
Note that the key assumption here is that the examples in the training set and the test objects are \textit{independent and identically distributed} (in fact, even a weaker requirement of \textit{exchangeability} is sufficient).

The idea is then to compute for a test object a $p$-value of every possible choice of the label.

Once the $p$-values are computed, they can be put to use in one of the following ways: 
\begin{itemize}
\item Given a significance level, $\epsilon$, a \textit{region predictor} outputs for each test object the set of labels (i.e., a region in the label space) such that the actual label is not in the set no more than a fraction $\epsilon$ of the times. 
If the prediction set consists of more than one label, the prediction is called {\it uncertain}, whereas
if there are no labels in the prediction set, the prediction is {\it empty}.
\item Alternatively, a {\it forced} prediction (chosen by the largest $p$-value)  is given, alongside with its \textit{credibility} (the largest $p$-value) and \textit{confidence} (the complement to 1 of the second largest $p$-value).
\end{itemize}

\subsection{Inductive Mondrian Conformal Predictors}

In order to apply conformal predictors to both big and imbalanced datasets,
we combine two variants of conformal predictors from~\cite{algoLearning,HedgingPreds}: Inductive (to reduce computational complexity) and Mondrian (to deal with imbalanced data sets) Conformal Predictors. 

To combine the Mondrian Conformal Prediction with that of Inductive Conformal Prediction, we have to revise the definition of $p$-value for the Mondrian case so that it incorporates the changes brought about by splitting the training set and evaluating the $\alpha_i$ only in the calibration set.

It is customary to split the training set at index $h$ so that examples with index $i \le h$ constitute the proper training set and examples with index $i > h$ (and $i \le \ell$) constitute the calibration set.

The $p$-values for a hypothesis $y_{\ell+1}=y$ about the label of $x_{\ell+1}$ are defined as
$$p(y)=\frac{|\{i=h+1,\dots,\ell+1:y_i=y,\alpha_i\ge\alpha_{\ell+1}\}|}{|\{i=h+1,\dots,\ell+1:y_i=y\}|}$$
In other words, the formula above considers only $\alpha_i$ associated with those examples in the calibration set that have the same label as that of the completion we are currently considering (note that also $\alpha_{\ell+1}$ is included in the set of $\alpha_i$ used for the comparison).
As in the previous forms of $p$-value, the fraction of such $\alpha_i$ that are greater than or equal to $\alpha_{\ell+1}$ is the $p$-value.

Finally, it is important to note that Inductive Conformal Predictors can be applied under less restrictive conditions. The requirement of i.i.d.\ can in fact be dropped for the proper training set, as the i.i.d.\ property is relevant only for the populations on which we calculate and compare the $\alpha_i$, that is, the calibration and testing set.

This will allow us to use NCM based on such methods as Cascade SVM (described in sec.\ref{sec:underlying}), including a stage of splitting big data into parts.

\section{Application to Compound Activity Prediction}
To evaluate the performance of CP for Compound Activity Prediction in a realistic scenario, we sourced the data sets from a public-domain repository of High Throughput assays, PubChem BioAssay~\cite{pubChem}. 

The data sets on PubChem identify a compound with its CID (a unique compound identifier that can be used to access the chemical data of the compound in another PubChem database) and provide the result of the assay as Active/Inactive as well as providing the actual measurements on which the result was derived, e.g. viability (percentage of cells alive) of the sample after exposure to the compound.

To apply machine learning techniques to this problem, the compounds must be described in terms of a number of numerical attributes. There are several approaches to do this. The one that was followed in this study is to compute \textit{signature descriptors}~\cite{FaulonSignature}. Each signature corresponds to the number of occurrences of a given labelled subgraph in the molecule graph, with subgraphs limited to those with a given depth. In this exercise the signature descriptors\footnote{The signature descriptors and other types of descriptors (e.g. circular descriptors) can be computed with the CDK Java package or any of its adaptations such as the RCDK package for the R statistical software.} had at most height 3.
Examples can be found in~\cite{LarsSLDS}.
The resulting data set is a sparse matrix of attributes (the signatures, on the columns) and examples (the compounds, on the rows).

We evaluated Conformal Predictors first with various underlying algorithms on the smallest of the data sets and then with various data sets using the underlying algorithm that performed best in previous set of tests.

\subsection{Underlying algorithms}
\label{sec:underlying}
As a first step in the study, we set out to extract relevant non-conformity measures from different underlying algorithms: Support Vector Machines (SVM), Nearest Neighbours, Na\"{i}ve Bayes. The Non Conformity Measures for each of the three underlying algorithms are listed in Table~\ref{tab:NCM}.

\begin{table}
\begin{center}
\caption{The Non Conformity Measures for the three underlying algorithms}
\label{tab:NCM}
\renewcommand{\arraystretch}{2}
\setlength{\tabcolsep}{1em} {
	\begin{tabular}{ |m{2cm}|m{3cm}|m{5cm}|}
	\hline
	\textbf{Underlying} & \textbf{Non Conformity Measure $\alpha_i$} & \textbf{Comment} \\
	\hline
	 SVM &     \(\displaystyle  -y_i d(x_i)\) & (signed) distance from separating hyperplane\\ 
	 \hline
	 kNN &            \(\displaystyle \frac{\sum^{(k)}_{j\neq i: y_j=y_i}d(x_j,x_i)}{\sum^{(k)}_{j\neq i: y_j \neq y_i}d(x_j,x_i)} \) & here the summation is on the $k$ smallest values of $d(x_j,x_i)$\\ 

	  \hline
	 Na\"{i}ve Bayes &    \(\displaystyle -{\log} \enspace p(y_i=c|x_i) \) & $p$ is the posterior probability estimated by Na\"{i}ve Bayes\\
	\hline
	\end{tabular}
	}
\end{center}

\end{table}


There are a number of considerations arising from the application of each of these algorithms to Compound Activity Prediction.

{\bf SVM.} The usage of SVM in this domain poses a number of challenges.
First of all, the number of training examples 
was large enough to create a problem for our computational resources. The scaling of SVM to large data sets is indeed an active research area~\cite{LargeScaleSVM,psvm,HybridMPIParSVM,SVMonHPC}, especially in the case of non-linear kernels\footnote{In the case of linear SVM, it is possible to tackle the formulation of the quadratic optimization problem at the heart of the SVM in the primal and solve it with techniques such as Stochastic Gradient Descent or L-BFGS, which lend themselves well to being distributed across an array of computational nodes}. We turned our attention to a simple approach proposed by V.Vapnik et al. in~\cite{cascadesvm}, called Cascade SVM. 

The sizes of the training sets considered here are too large to be handled comfortably by generally available SVM implementations, such as \verb|libsvm|~\cite{libsvm}. 
The approach we follow could be construed as a form of \textit{training set editing}.
Vapnik proved formally that it is possible to decompose the training into an $n$-ary tree of SVM trainings. The first layer of SVMs is trained on training sets obtained as a partition of the overall training set. Each SVM in the first layer outputs its set of support vectors (SVs) which is generally smaller than the training set. In the second layer, each SVM takes as training set the merging of $n$ of the SVs sets found in the first layer. Each layer requires fewer SVMs. The process is repeated until a layer requires only one SVM. The set of SVs emerging from the last layer is not necessarily the same that would be obtained by training on the whole set (but it is often a good approximation).
If one wants to obtain that set, the whole training tree should be executed again, but this time the SVs obtained at the last layer would be merged into each of the initial training blocks. A new set of SVs would then be obtained at the end of the tree of SVMs. If this new set is the same as the one in the previous iteration, this is the desired set. If not, the process is repeated once more.
In~\cite{cascadesvm} it was proved that the process converges and that it converges to the same set of SVs that one would obtain by training on the whole training set in one go.

To give an intuitive justification, the fundamental observation is that the SVM decision function is entirely defined just by the Support Vectors. It is as if these examples contained all the information necessary for the classification. Moreover, if we had a training set composed only of the SVs, we would have obtained the same decision function. So, one might as well remove the non-SVs altogether from the training set.

\begin{figure}[htpb]
\begin{center}
\includegraphics[width=0.5\textwidth]{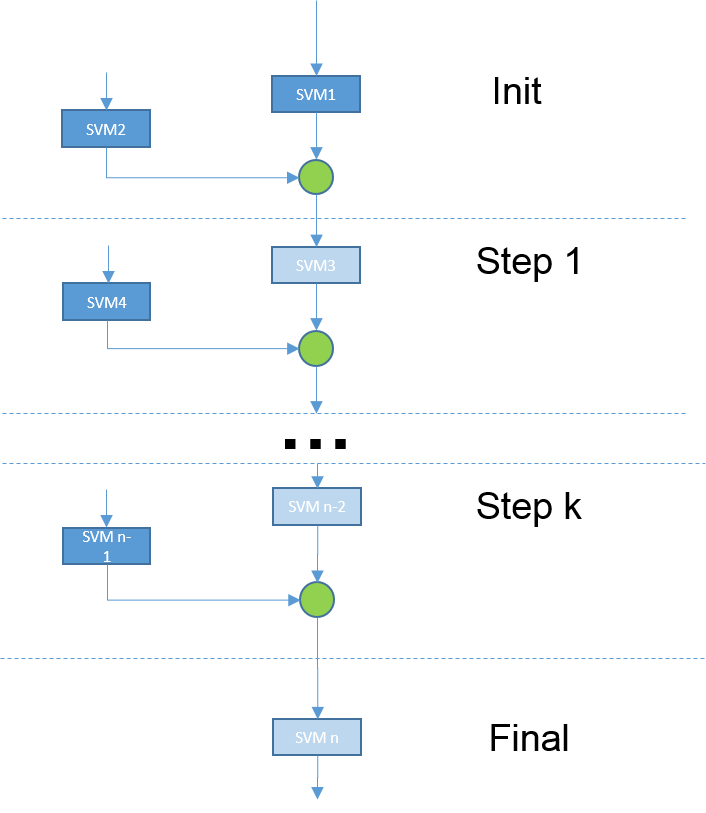} 
\caption[]{Linear Cascade SVM \\
At each step, the set of Support Vectors from the previous stage is merged with a block of training examples from the partition of the original training set. This is used as training set for an SVM, whose SVs are then fed to the next stage.}
\label{linCascade}
\end{center}
\end{figure}

In experiments discussed here, we followed a simplified approach. Instead of a tree of SVMs, we opted for a linear arrangement as shown in Fig.\ref{linCascade}. 

While we have no theoretical support for this semi-online variant of the Cascade SVM, the method appears to work satisfactorily in practice on the data sets we used.

The class imbalance was addressed with the use of per-class weighting of the $C$ hyperparameter, which results in a different penalization of the margin violations. The per-class weight was set inversely proportional to the class representation in the training set.

Another problem is the choice of an appropriate kernel.
While we appreciated the computational advantages of linear SVM, we also believed that it was not necessarily the best choice for the specific problem. It can easily be observed that the nature of the representation of the training objects (as discrete features) warranted approaches similar to those used in Information Retrieval, where objects are described in terms of occurrences of patterns (bags of words). The topic of similarity searching in chemistry is an active one and there are many alternative proposals (see \cite{similaritySearching}). We used as a kernel a notion called Tanimoto similarity\footnote{See \cite{kernelsStructured} for a proof that Tanimoto Similarity is a kernel.}. The Tanimoto similarity extends the well-known Jaccard coefficient in the sense that whereas the Jaccard coefficient considers only presence or absence of a pattern, the Tanimoto similarity takes into account the counts of the occurrences.

To explore further the benefits of non-linear kernels, we also tried out a kernel consisting of the composition the Tanimoto similarity with Gaussian RBF.

Table~\ref{tab:kernels} provides the definitions of the kernels used in this study.

\begin{table}[htpb]
\caption{SVM Kernels Definitions (where $A=(a_1,\dots,a_d),B=(b_1,\dots,b_d)$ are two objects, each described by a vector of $d$ counts) )}
\label{tab:kernels}
\begin{center}
$$
\begin{array}{|m{5cm}|m{7cm}|}
\hline
 Tanimoto Coefficient &   \begin{center}$ {T(A,B) = \frac{\sum_{i=1}^d \min(a_i,b_i)}{\sum_{i=1}^d (a_i + b_i) - \sum_{i=1}^d \min(a_i,b_i)}} $ \end{center} \\ \hline
 Tanimoto with Gaussian RBF & \begin{center} $  TG(A, B)=e^{-\frac{\left|{T(A,A)+T(B,B)-2T(A,B)}\right|}{\gamma}}$ \end{center} \\   \hline
\end{array}
$$
\end{center}
\end{table}

\subsubsection{Na\"{i}ve Bayes.}

Na\"{i}ve Bayes and more specifically Multinomial Na\"{i}ve Bayes are widely regarded as effective classifiers when features are discrete (for instance, in text classification), despite their relative simplicity. This made Multinomial Na\"{i}ve Bayes a natural choice for the problem at issue here.

In addition, Na\"{i}ve Bayes has a potential of providing some guidance for feature selection, via the computed posterior probabilities. This is of particular interest in the domain of Compound Activity Prediction, as it may provide insight as to the molecular structures that are associated with Activity in a given assay. This knowledge could steer further testing in the direction of a class of compounds with higher probability of Activity.

\subsubsection{Nearest Neighbours.}
We chose Nearest Neighbours because of its good performance in a wide variety of domains.
In principle, the performance of Nearest Neighbours could be severely affected by the high-dimensionality of the training set (Talble~\ref{tab:aid827} shows how in one of the data sets used in this study the number of attributes exceeds by $\approx20\%$ the number of examples), but some preliminary small-scale 
experiments did not show that this causes the ``curse of dimensionality''. 

\subsection{Tools and Computational Resources}
The choice of the tools for these experiments was influenced primarily by the exploratory nature of this work. For this reason, tools, programming languages and environments that support interactivity and rapid prototyping were preferred to those that enable optimal CPU and memory efficiency.

The language adopted was Python 3.4 and the majority of programming was done using IPython Notebooks in the Jupyter environment. 
The overall format turned out to be very effective for capturing results (and for their future reproducibility).

Several third-party libraries were used. 
The computations were run initially on a 
local server (8 cores with 32GB of RAM, running OpenSuSE) and in later stages on a supercomputer (the IT4I Salomon cluster located in Ostrava, Czech Republic).
The Salomon cluster is based on the SGI® ICE™ X system and comprises 1008 computational nodes (plus a number of login nodes), each with 24 cores (2 12-core Intel Xeon E5-2680v3 2.5GHz processors) and 128GB RAM, connected via high-speed 7D Enhanced hypercube InfiniBand FDR and Ethernet networks. It currently ranks at \#48 in the top500.org list of supercomputers and at \#14 in Europe\footnote{According to
 \path{https://www.sgi.com/company_info/newsroom/press_releases/2015/september/salomon.html}.}.

Parallelization and computation distribution relied on the \verb|ipyparallel|\cite{ipyparallel} package, which is a high-level framework for the coordination of remote execution of Python functions on a generic collection of nodes (cores or separate servers). While \verb|ipyparallel| may not be highly optimized, it aims at providing a convenient environment for distributed computing well integrated with IPython and Jupyter and has a learning curve that is not as steep as that of the alternative frameworks common in High Performance Computing (OpenMPI, for example).
In particular, \verb|ipyparallel|, in addition to allowing the start-up and shut-down of a cluster comprising a controller and a number of engines where the actual processing (each is a separate process running a Python interpreter) is performed via integration with the job scheduling infrastructure present on Salomon (PBS, Portable Batch System), took care of the details such as data serialization/deserialization and transfer, load balancing, job tracking, exception propagation, etc. thereby hiding much of the complexity of parallelization.
One key characteristic of \verb|ipyparallel| is that, while it provides primitives for \verb|map()| and \verb|reduce()|, it does not constrain the choice to those two, leaving the implementer free to select the most appropriate parallel programming design patterns for the specific problem (see ~\cite{structuredParallelProgramming} for a reference on the subject).

In this work, parallelization was exploited to speed up the computation of the Gram matrix or of the decision function for the SVMs or the matrix of distances for kNN. In either case, the overall task was partitioned in smaller chunks that were then assigned to engines, which would then asynchronously return the result. Also, parallelization was used for SVM cross-validation, but at a coarser granularity, i.e. one engine per SVM training with a parameter. Data transfers were minimized by making use of shared memory where possible and appropriate. A key speed-up was achieved by using pre-computed kernels (computed once only) when performing Cross-Validation with respect to the hyperparameter C.

\subsection{Results}
To assess the relative merits of the different underlying algorithms, we applied Inductive Mondrian Conformal Predictors on data set AID827, whose characteristics are listed in Table~\ref{tab:aid827}.

\begin{table}[htpb]
\caption{Characteristics of the AID827 data set }
\label{tab:aid827}
\begin{center}
		\setlength{\tabcolsep}{0.2cm}
		\begin{tabular}{lrl}
		Total number of examples  & 138,287\\
		Number of features  & 165,786 & High dimensionality\\
		Number of non-zero entries  & 7,711,571\\
		Density of the data set & 0.034\% & High sparsity\\
		Active compounds  & 1,658 & High imbalance (1.2\%)\\
		Inactive compounds  &  136,629 \\
		Unique set of signatures  &  137,901 & Low degeneracy \\
		\end{tabular}

\end{center}
\end{table}

The test was articulated in 20 cycles of training and evaluation. In each cycle, a test set of 10,000 examples was extracted at random. The remaining examples were split randomly into a proper training set of 100,000 examples and a calibration set with the balance of the examples (28,387).

During the SVM training, 5-fold stratified Cross Validation was performed at every stage of the Cascade to select an optimal value for the hyperparameter C. 
Also, per-class weights were assigned to cater for the high class imbalance in the data, so that a higher penalization was applied to violators in the less represented class.

In Multinomial Na\"{i}ve Bayes too, Cross Validation was used to choose an optimal value for the smoothing parameter.

The results are listed in Table~\ref{tab:827CP}, which presents the classification arising from the region predictor for $\epsilon=0.01$. The numbers are averages over the 20 cycles of training and testing.

Note that a compound is classified as Active (resp. Inactive) if and only if Active (resp. Inactive) is the only label in the prediction set. When both labels are in the prediction, the prediction is considered Uncertain.

\begin{table}[htpb]
\caption{CP results for AID827 with significance $\epsilon=0.01$. All results are averages over 20 runs, using the same test sets of 10,000 objects across the different underlying algorithms. ``Active predicted Active'' is the (average) count of actually Active test examples that were predicted Active by Conformal Prediction. Uncertain predictions occur when both labels are output by the region predictor. Empty predictions occur when both labels can be rejected at the chosen significance level.
For the specific significance level chosen here, there were never empty predictions.}
\label{tab:827CP}
\begin{center}
	\begin{tabular}{l R{1.1cm} R{1.2cm} R{1.3cm} R{1.3cm} R{1cm} R{1.2cm}}
	Underlying & Active pred Active & Inactive pred Active & Inactive pred Inactive & Active pred Inactive & Empty  pred & Uncertain \\
	\hline
	Na\"{i}ve Bayes & 38.20 & 104.30 & 183.30 & 1.10 & 0 & 9673.10 \\\hline
	3NN & 43.95 & 100.55 & 361.55 & 0.80& 0 & 9493.15 \\\hline 	
          Cascade SVM: & & & & & & \\
	- linear & 34.20 & 99.00 & 591.85 & 1.20 & 0 & 9273.75 \\
	- RBF kernel & 47.20 & 101.80 & 1126.75 & 1.80 & 0 & 8722.45 \\
	- Tanimoto kernel& 48.45 & 97.65 & 986.85 & 0.80 & 0 & 8866.25 \\
	- Tanimoto-RBF kernel& 47.65 & 94.10 & 1044.90 & 0.95 & 0 & 8812.40 \\\hline
	\end{tabular}
\end{center}
\end{table}

It has to be noted at this stage that there does not seem to be an established consensus on what the best performance criteria are in the domain of Compound Activity Prediction (see for instance~\cite{recommEval}), although \textit{Precision} (fraction of actual Actives among compounds predicted as Active) and \textit{Recall} (fraction of all the Active compounds that are among those predicted as Active) seem to be generally relevant. In addition, it is worth pointing out that these (and many others) criteria of performance should be considered as generalisations of classical performance criteria since they include dependence of the results on the required confidence level.

\begin{table}[htpb]
\caption{CP results for AID827 using SVM with Tanimoto+RBF kernel for different significance levels. The ``Active Error Rate'' is the ratio of ``Active predicted Inactive'' to the total number of Active test examples.  The ``Inactive Error Rate'' is the ratio of ``Inactive predicted Active'' to the total number of Inactive test examples.}
\label{tab:827CPVarEps}
\begin{center}
	\begin{tabular}{l R{1.1cm} R{1.2cm} R{1.3cm} R{1.3cm} R{1cm} R{1.2cm} R{1.2cm} R{1.2cm}}
	Significance & Active pred Active & Inactive pred Active & Inactive pred Inactive & Active pred Inactive & Empty  pred & Uncertain & Active Error Rate & Inactive Error Rate\\

	\hline
	1\% &       47.65 &         94.10 &       1044.90 &        0.95 &        0.0 &      8812.40  &  0.82\% & 0.95\%\\
	5\% &       67.20 &        490.40 &       3091.75 &        5.20 &        0.0 &      6345.45 &  4.52\% & 4.96\%\\
	10\% &       76.15 &        999.25 &       4703.75 &       10.60 &        0.0 &      4210.25 &  9.22\% & 10.11\%\\
	15\% &       82.10 &       1484.85 &       6021.80 &       17.30 &        0.0 &      2393.95 &  15.04\% & 15.02\%\\
	20\% &       86.55 &       1982.25 &       6928.95 &       22.80 &        0.0 &       979.45 &  19.83\% & 20.05\%\\
	
	\end{tabular}
\end{center}
\end{table}
At the shown significance level of $\epsilon=0.01$, $34\%$ of the compounds predicted as active by Inductive Mondrian Conformal Prediction using Tanimoto composed with Gaussian RBF were actually Active compared to a prevalence of Actives in the data set of just $1.2\%$.  
At the same time, the Recall was $\approx 41\%$ 
(ratio of Actives in the prediction to total Actives in the test set).

We selected Cascade SVM with Tanimoto+RBF as the most promising underlying algorithm on the basis of the combination of its high Recall (for Actives) and high Precision (for Actives), assuming that the intended application is indeed to output a selection of compounds that has a high prevalence of Active compounds.

Note that in Table~\ref{tab:827CP} the values similar to ones of confusion matrix are calculated only for certain predictions.
In this representation, the concrete meaning of the property of class-based validity can be clearly illustrated as in Table~\ref{tab:827CPVarEps}: the two rightmost columns report the prediction error rate for each label, where by prediction error we mean the occurrence of ``the actual label not being in the predictions set''. When there are no Empty predictions, the Active Error rate is the ratio of the number of ``Active predicted Inactive'' to the number of Active examples in the test set (which was 115 on average). 

\begin{figure}[htpb]
\caption{Test objects plotted by the base-10 log of their their $p_{active}$ and $p_{inactive}$. Note that many test objects are overlapping. 
Note that some of the examples may have identical p-values, so for example 1135 objects predicted as "Inactives" are presented as 4 points on this plot.
}
\label{fig:log10p}
\begin{center}

\includegraphics[width=0.85\textwidth]{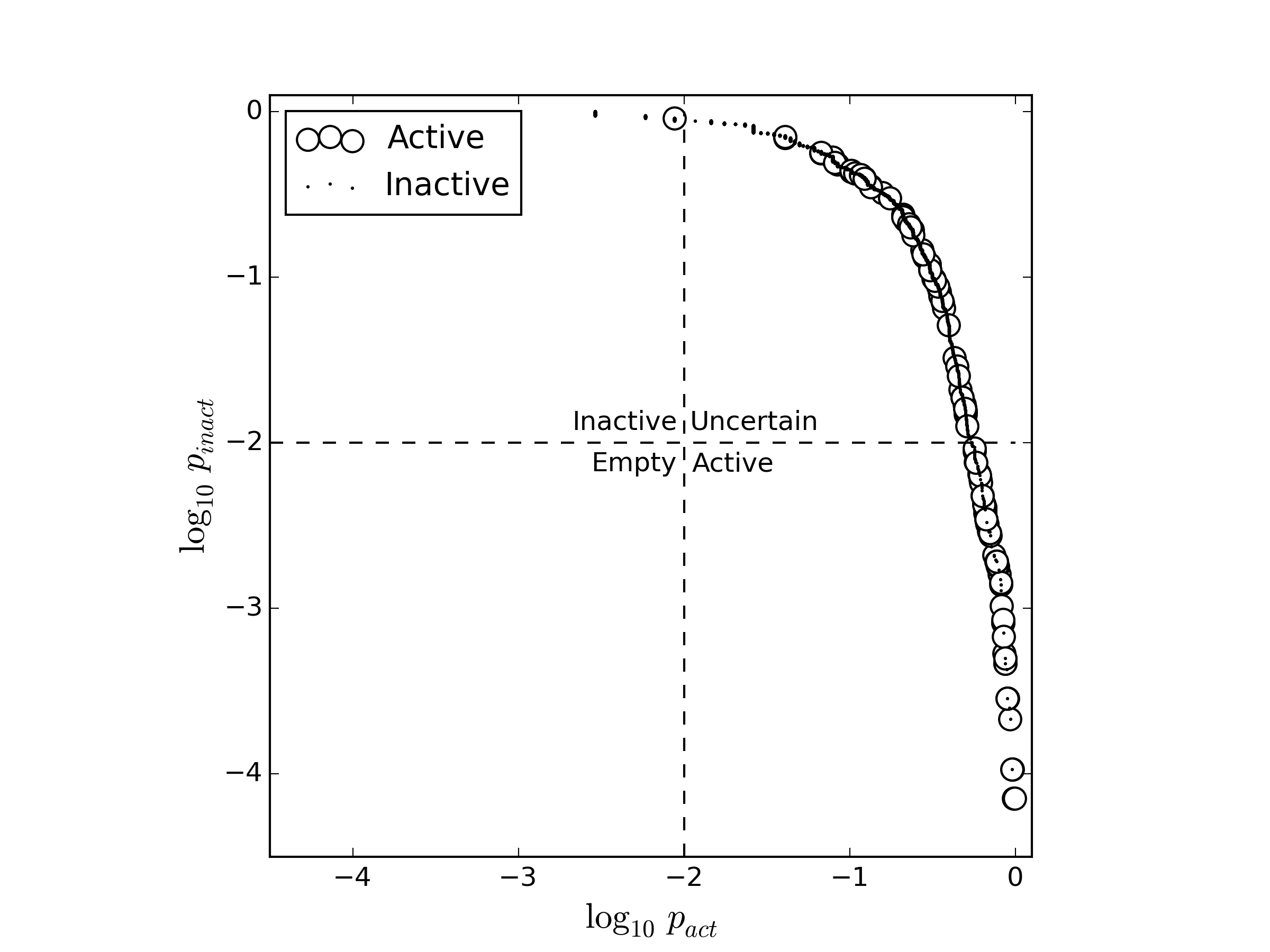}

\end{center}
\end{figure}

Fig.~\ref{fig:log10p} shows the test objects according to the base-10 logarithm of their $p_{active}$ and $p_{inactive}$.
The dashed lines represent the thresholds for $p$-value set at 0.01, i.e. the significance value $\epsilon$ used in Table~\ref{tab:827CP}. The two dashed lines partition the plane in 4 regions, corresponding to the region prediction being Active ($p_{active} > \epsilon$ and $p_{inactive} \leq \epsilon$), Inactive ($p_{active} \leq \epsilon$ and $p_{inactive} > \epsilon$), Empty ($p_{active} \leq \epsilon$ and $p_{inactive} \leq \epsilon$), Uncertain ($p_{active} > \epsilon$ and $p_{inactive} > \epsilon$).

As we said in Sec.~\ref{cp01}, the alternative is forced prediction with individual confidence and credibility.

It is clear that there are several benefits accruing from using Conformal Predictors.
For instance, a high $p$-value for the Active hypothesis might suggest that Activity cannot be ruled out,  but the same compound may exhibit also a high $p$-value for the Inactive hypothesis, which would actually mean that neither hypothesis could be discounted.


In this specific context it can be argued that the $p$-values for Active hypothesis are more important. They can be used to rank the test compounds like it was done in~\cite{ppi} for ranking potential interaction.
A high $p$-value for the Active hypothesis might suggest that Activity cannot be ruled out. For example it is possible to output the prediction list of all compounds with $p$-values above a threshold $\epsilon=0.01$. 
A concrete activity which is not yet discovered will be covered by this list with probability $0.99$.
All the rest examples are classified as Non-Active with confidence $0.99$ or larger.


Special attention should be also paid to {\it low credibility} examples where both $p$-values are small. Intuitively, low credibility means that
either the training set is non-random
or the test object is not representative of the training set. 
For such examples, the label assignment does 
not conform to the training data.. They may be considered as anomalies or examples of compound types not enough represented in the training set. This may suggest that it would be beneficial to the overall performance of the classifier to perform a lab test for those compounds and include the results in training set. Typically credibility will not be low
provided the data set was generated independently from the same distribution:
the probability that credibility
will be less than some threshold $\epsilon$ (such as 1\%)
is less than $\epsilon$.

\medskip

\begin{figure}[htpb]
\caption{Trade-off between Precision and Recall by varying Credibility or Confidence}
\label{fig:varyCredibility}
\begin{center}
{
\includegraphics[width=0.45\textwidth]{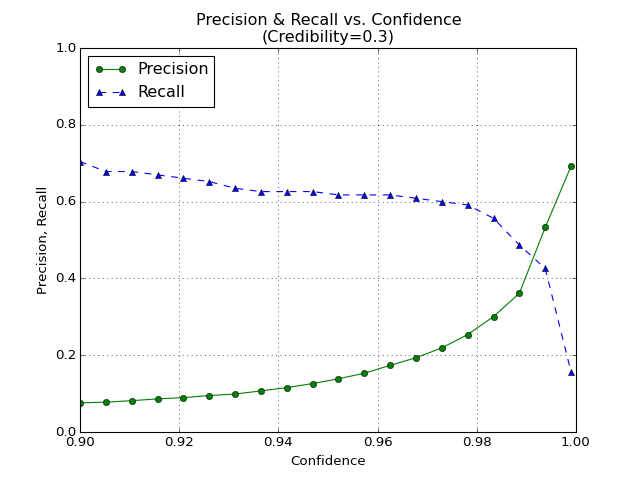}
\includegraphics[width=0.45\textwidth]{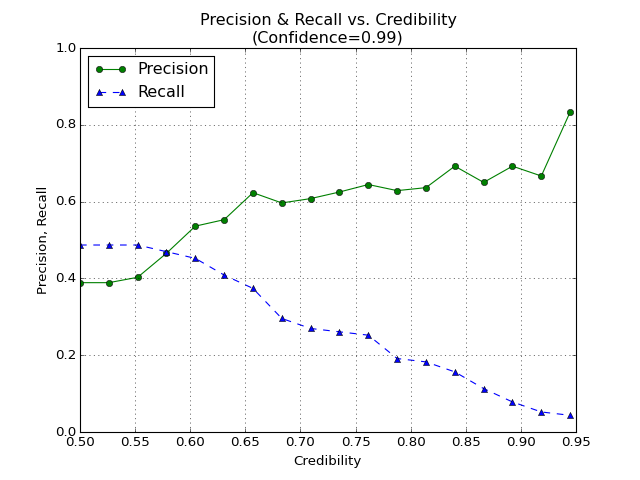}
}
\end{center}
\end{figure}

Finally, Conformal Predictors provide the user with the additional degree of freedom of the significance or confidence level. By varying either of those two parameters, a different trade-off between Precision and Recall or any of the other metrics that are of interest can be chosen.
Fig.~\ref{fig:varyCredibility} illustrates this point with two examples. 
The Precision and Recall shown in the two panes were calculated on the test examples predicted Active which exceeded both a Credibility threshold and a given the Confidence threshold. In the left pane, the Credibility threshold was fixed and the Confidence threshold was varied; vice versa in the right pane.

\subsection{Application to different data sets}
We applied Inductive Conformal Predictors with underlying SVM using Tanimoto+RBF kernel to other data sets extracted from PubChem BioAssay to verify if the same performance would be achieved for assays covering a range of quite different biological targets and to what extent the performance would vary with differences in training set size, imbalance, and sparseness of the training set. The main characteristics of the data sets are reported in Table~\ref{tab:DataSets}.

As in the previous set of experiments, 20 cycles of training and testing were performed and the results averaged over them. In each cycle, a test set of 10,000 examples was set aside and the rest was split between calibration set ($\approx30,000$) and proper training set.
The results are reported in Table~\ref{tab:multiDataSetsRes}.

It can be seen that five data sets differ in their hardness for machine learning some of the produce more uncertain predictions using the same algorithms, number of examples and the same significance level.

\begin{table}[htpb]
\caption{Data sets and their characteristics. Density refers to the percentage of non-zero entries in the full matrix of `Number of Compounds $\times$ Number of Features' elements}
\label{tab:DataSets}
\begin{center}
	\renewcommand{\arraystretch}{2}
	\begin{tabular}{m{0.8cm} 
	                m{5cm} 
	                >{\centering\arraybackslash}m{2.2cm} 
	                >{\centering\arraybackslash}m{1.6cm} 
	                >{\centering\arraybackslash}m{1.3cm} 
	                >{\centering\arraybackslash}m{1cm}}
	Data Set & Assay Description & Number of Compounds & Number of Features  & Actives (\%) & Density (\%) \\
	\hline
	827 & High Throughput Screen to Identify Compounds that Suppress the Growth of Cells with a Deletion of the PTEN Tumor Suppressor.  & 138,287 & 165,786 & 1.2\% & 0.034\% \\
	1461 & qHTS Assay for Antagonists of the Neuropeptide S Receptor: cAMP Signal Transduction.  & 208,069 & 211,474 & 1.11\% & 0.026\%  \\
	1974 & Fluorescence polarization-based counterscreen for RBBP9 inhibitors: primary biochemical high throughput screening assay to identify inhibitors of the oxidoreductase glutathione S-transferase omega 1(GSTO1). & 302,310 & 237,837 & 1.05\% & 0.024\% \\
	2553 & High throughput screening of inhibitors of transient receptor potential cation channel C6 (TRPC6) & 305,308 & 236,508 & 1.06\% & 0.024\%  \\
	2716 & Luminescence Microorganism Primary HTS to Identify Inhibitors of the SUMOylation Pathway Using a Temperature Sensitive Growth Reversal Mutant Mot1-301 & 298,996 & 237,811 & 1.02\% & 0.024\% \\
	\end{tabular}
\end{center}
\end{table}

\begin{table}[htpb]
\caption{Results of the application of Mondrian ICP with $\epsilon=0.01$ using SVM with Tanimoto+RBF as underlying. Test set size: 10,000}
\label{tab:multiDataSetsRes}
\begin{center}
\setlength{\tabcolsep}{0.3cm}
\begin{tabular}{l R{1.1cm} R{1.2cm} R{1.3cm} R{1.3cm} R{1cm} R{1.2cm}}
\toprule
DataSet &  Active pred Active & Inactive pred Active & Inactive pred Inactive & Active pred Inactive & Empty  pred & Uncertain \\
\midrule
827     &       47.65 &          94.10 &       1044.90 &        0.95 &          0 &      8812.40 \\
1461    &       29.45 &         101.30 &       1891.10 &        1.20 &          0 &      7976.95 \\
1974    &       62.50 &          97.40 &        880.85 &        1.00 &          0 &      8958.25 \\
2553    &       34.00 &         101.00 &        337.90 &        1.00 &          0 &      9526.10 \\
2716    &        3.55 &          98.20 &         97.00 &        1.00 &          0 &      9800.25 \\
\bottomrule
\end{tabular}

\end{center}
\end{table}

\subsection{Mondrian ICP with different $\epsilon_{active}$ and $\epsilon_{inactive}$}

When applying 
Mondrian ICP, there is no constraint to use the same significance $\epsilon$ for the two labels. There may be an advantage in allowing different ``error'' rates for the two labels given that the focus might be in identifying Actives rather than Inactives.

This allows to vary relative importance of the two kinds of errors.
Validity of Mondrian machines implies that the expected number of certain but wrong predictions is bounded by $\epsilon_{act}$ for (true) actives
and by $\epsilon_{inact}$ for (true) non-actives. 
It is interesting to study its effect also on the precision and recall (within certain prediction).

\begin{figure}[htpb]
\caption{Trade-off between Precision and Recall by varying $\epsilon_{inact}$}
\label{fig:varyEpsInact}
\begin{center}
\includegraphics[width=0.8\textwidth]{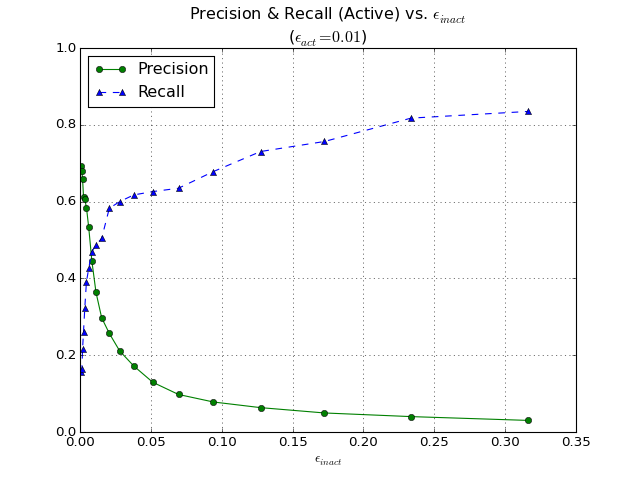}
\end{center}
\end{figure}

Fig.~\ref{fig:varyEpsInact} shows the trade-off between Precision and Recall that results from varying $\epsilon_{inact}$.

For very low values of the significance $\epsilon$, a large number of test examples have $p_{act}>\epsilon_{act}$ as well as $p_{inact}>\epsilon_{inact}$. For these test examples, we have an `Uncertain' prediction.

As we increase $\epsilon_{inact}$, fewer examples have a $p_{inact}$ larger than $\epsilon_{inact}$. So `Inactive' is not chosen any longer as a label for those examples. If they happen to have a $p_{act} > \epsilon_{act}$, they switch from 'Uncertain' to being predicted as `Active' (in the other case, they would become `Empty' predictions).

\begin{figure}[htpb]
\caption{Precision vs. Recall: three methods}
\label{fig:PrecisionVsRecall}
\begin{center}
\includegraphics[width=0.8\textwidth]{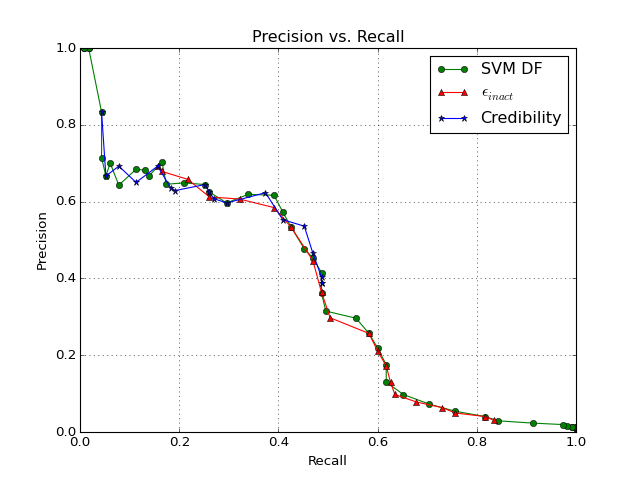}
\end{center}
\end{figure}

Fig.~\ref{fig:PrecisionVsRecall} shows how Precision varies with Recall using three methods: varying the threshold applied to the Decision Function of the underlying SVM, varying the significance $\epsilon_{inact}$ for the Inactive class, varying the credibility.
The three methods give similar results.

\section{Conclusions}

This paper summarized a methodology of applying conformal prediction to big and imbalanced data  with several underlying methods like nearest neighbours, Bayes, SVM and various kernels. The results have been compared from the point of view of efficiency of various methods and various sizes of the data sets.


The paper also presents results of using Inductive Mondrian Conformal Predictors with different significance levels for different classes.

The most interesting direction of the future extension is to study the possible strategies of active learning (or experimental design). In this paper one of the criteria of performance is a number of uncertain predictions. It might be useful to select among them the compounds that should be checked experimentally first -- in other words  the most ``promising'' compounds. How to select though may depend on practical scenarios of further learning and on comparative efficiency of different active learning strategies.


\section{Acknowledgments}

This project (ExCAPE) has received funding from the European Union’s Horizon 2020 Research and Innovation programme under Grant Agreement no. 671555. 
We are grateful for the help in conducting experiments to the Ministry of Education, Youth and Sports (Czech Republic) that supports the Large Infrastructures for Research, Experimental Development and Innovations project ``IT4Innovations National Supercomputing Center – LM2015070''. This work was also supported by
EPSRC grant EP/K033344/1 (``Mining the Network Behaviour of Bots''). 
We are indebted to Lars Carlsson of Astra Zeneca for providing the data and useful discussions. We are also thankful to Zhiyuan Luo and Vladimir Vovk for many valuable comments and discussions.

%
%

\end{document}